\def\year{2022}\relax
\documentclass[letterpaper]{article} 
\usepackage{aaai22}  
\usepackage{times}  
\usepackage{helvet}  
\usepackage{courier}  
\usepackage[hyphens]{url}  
\usepackage{graphicx} 
\usepackage{natbib}  
\usepackage{caption} 
\DeclareCaptionStyle{ruled}{labelfont=normalfont,labelsep=colon,strut=off} 
\usepackage{newfloat}
\usepackage{listings}
\usepackage{pdfpages}
\usepackage{amsmath}
\usepackage{amssymb}
\usepackage{mathtools}
\usepackage{amsthm}
\usepackage{mathrsfs}
\usepackage{enumitem}
\usepackage{subfig}
\usepackage{multirow}
\usepackage{xcolor}         
\usepackage{booktabs}
\usepackage[linesnumbered,boxed,ruled]{algorithm2e}

\DeclareMathOperator*{\pr}{\text{Pr}}
\DeclareMathOperator*{\ex}{\mathbb{E}}

\newtheorem{theorem}{Theorem}
\newtheorem{lemma}{Lemma}
\newtheorem{definition}{Definition}

\title{Learning by Competition of Self-Interested Reinforcement Learning Agents}
\author {Stephen Chung}
\affiliations{    
	Department of Computer Science, University of Massachusetts Amherst, USA\\
	minghaychung@umass.edu 
}

\begin{document}	
	\maketitle	
	\begin{abstract}
		An artificial neural network can be trained by uniformly broadcasting a reward signal to units that implement a REINFORCE learning rule. Though this presents a biologically plausible alternative to backpropagation in training a network, the high variance associated with it renders it impractical to train deep networks. The high variance arises from the inefficient structural credit assignment since a single reward signal is used to evaluate the collective action of all units. To facilitate structural credit assignment, we propose replacing the reward signal to hidden units with the change in the $L^2$ norm of the unit's outgoing weight. As such, each hidden unit in the network is trying to maximize the norm of its outgoing weight instead of the global reward, and thus we call this learning method \emph{Weight Maximization}. We prove that Weight Maximization is approximately following the gradient of rewards in expectation. In contrast to backpropagation, Weight Maximization can be used to train both continuous-valued and discrete-valued units. Moreover, Weight Maximization solves several major issues of backpropagation relating to biological plausibility. Our experiments show that a network trained with Weight Maximization can learn significantly faster than REINFORCE and slightly slower than backpropagation. Weight Maximization illustrates an example of cooperative behavior automatically arising from a population of self-interested agents in a competitive game without any central coordination.
	\end{abstract}
	
	\section{Introduction}
	The \emph{error backpropagation algorithm} (backprop) \cite{rumelhart1986learning} efficiently computes the gradient of an objective function with respect to parameters by iterating backward from the last layer of a multi-layer artificial neural network (ANN). However, backprop is generally regarded as being biologically implausible \cite{crick1989recent, mazzoni1991more, o1996biologically, bengio2015towards, hassabis2017neuroscience, lillicrap2020backpropagation}. First, the learning rule given by backprop is non-local, as it relies on information other than input and output of a neuron-like unit computed in the feedforward phase. Second, backprop requires synaptic symmetry in the forward and backward paths, which has not been observed in biological systems. Third, backprop requires precise coordination between the feedforward and feedback phase because the feedforward value has to be retained until the error signal arrives.
	
	Alternatively, REINFORCE \cite{williams1992simple}, a special case of $A_{R- \lambda P}$ when $\lambda$ = 0 \cite{barto1985pattern}, could be applied to all units as a more biologically plausible way of training a network. It is shown that the learning rule gives an unbiased estimate of the gradient of return \cite{williams1992simple}. Another interpretation of this relates to viewing each unit as a reinforcement learning (RL) agent, with each agent trying to maximize the same reward from the environment. We can thus view an ANN as a \emph{team of agents} playing a cooperative game, a scenario where all agents receive the same reward. Such a team of agents is also known as \emph{coagent network} \cite{thomas2011policy}. However, coagent networks can only feasibly solve simple tasks due to the high variance associated with this training method and thus the low speed of learning. The high variance stems from the lack of structural credit assignment, i.e.\ a single reward signal is used to evaluate the collective action of all agents.
	
	To address the lack of structural credit assignment in a team of agents trained by REINFORCE, we consider delivering a different reward signal to each hidden agent instead of the same global reward. Here hidden agents correspond to hidden units in an ANN and refer to the agents that output to other agents in the team instead of to the environment. As such, each hidden agent is associated with an \emph{outgoing weight}, that is, the vector of the weight by which the agent’s actions influence other agents in the team. We propose to replace the global reward signal to each hidden agent with the change in the $L^2$ norm of its outgoing weight, such that each hidden agent in the team is trying to maximize the norm of its outgoing weight. We call this new learning method \emph{Weight Maximization}. This is based on the intuition that the norm of an agent's outgoing weight roughly reflects the contribution of the agent in the team. This change of reward signals turns the original cooperative game into a competitive game since agents no longer receive the same reward. 
	
	We prove that Weight Maximization is approximately following the gradient of return in expectation, showing that every hidden agent maximizing the norm of its outgoing weight  also approximately maximizes the team's rewards. This illustrates an example of cooperative behavior automatically arising from a population of self-interested agents in a competitive game and offers an alternative perspective of training an ANN - each unit maximizing the norm of its outgoing weight instead of a network maximizing its performance. This alternative perspective localizes the optimization problem for each unit, yielding a wide range of RL solutions in training a network. Our experiments show that a network trained with Weight Maximization can learn much faster than REINFORCE, such that its speed is slightly lower than backprop.
	
	One may question whether the change of synaptic strength, analogous to the change of weights in ANNs, can be used to guide plasticity in biological systems, since the change of synaptic strength has a much slower timescale compared to the activation of neurons. To address this issue, we generalize Weight Maximization to use eligibility traces, such that the network can still learn when the outgoing weights change slowly. Weight Maximization with eligibility traces also solves the three aforementioned problems of backprop regarding biological plausibility. Nonetheless, the biological plausibility of Weight Maximization remains to be investigated. It is not yet clear if there exists a molecular mechanism that uses the change of synaptic strength in axons to guide the change of synaptic strength in dendrites.
	
	In summary, our paper has the following main contributions:
	
	\begin{itemize}
		\item We propose a novel algorithm called Weight Maximization that allows efficient structural credit assignment and significantly lowers the variance associated with REINFORCE when training a team of agents;  
		\item We prove that Weight Maximization is approximately following the gradient of return in expectation, establishing the approximate equivalence of hidden units maximizing the norm of outgoing weights and external rewards, thus providing theoretical justification for algorithms \cite{uhr1961pattern, klopf1969evolutionary, anderson1986learning, selfridge1988pandemonium} based on the norm of outgoing weights; 
		\item Weight Maximization can be used to train continuous-valued and discrete-valued units, offering an advantage to backprop;
		\item Weight Maximization represents a feasible alternative to backprop given their comparable learning speed; 
		\item We generalize Weight Maximization to use eligibility traces, which solves several major issues of backprop relating to biological plausibility.
	\end{itemize}
	
	The paper and the appendix is available at \url{https://arxiv.org/abs/2010.09770}.
	
	\section{Notation} \label{sec:n}
	
	We consider a Markov Decision Process (MDP) defined by a tuple $(\mathcal{S}, \mathcal{A}, P, R, \gamma, d_0)$, where $\mathcal{S}$ is a finite set of states of an agent's environment (although this work can be extended to the infinite state case), $\mathcal{A}$ is a finite set of actions, $P:\mathcal{S}\times \mathcal{A}\times \mathcal{S} \rightarrow [0,1]$ is a transition function giving the dynamics of the environment, $R: \mathcal{S}\times \mathcal{A} \rightarrow \mathbb{R}$ is a reward function, $\gamma \in [0,1]$ is a discount factor, and $d_0: \mathcal{S} \rightarrow [0,1]$ is an initial state distribution.  Denoting the state, action, and reward signal at time $t$ by  $S_t$, $A_t$, and $R_t$ respectively, $P(s, a, s') = \pr(S_{t+1}=s'|S_t=s, A_t=a)$, $R(s, a) = \ex[R_t|S_t=s, A_t=a]$, and $d_0(s) = \pr(S_{0}=s)$, where $P$ and $d_0$ are probability mass functions. An episode is a sequence of states, actions, and rewards, starting from $t=0$ and continuing until reaching the terminal state. For any learning methods, we can measure its performance as it improves over multiple episodes, which makes up a run.
	
	Letting $G_t = \sum_{k=t}^{\infty} \gamma^{k-t} R_k$ denote the infinite-horizon discounted return accrued after acting at time $t$, we are interested in finding, or approximating, a \emph{policy} $\pi:  \mathcal{S}\times \mathcal{A} \rightarrow [0,1]$ such that for any time $t > 0$, selecting actions according to $\pi(s,a)=\pr(A_t=a|S_t=s)$ maximizes the expected return $\ex[G_t]$. The value function for policy $\pi$ is $V^{\pi}$ where for all $s \in  \mathcal{S}$, $V^{\pi}(s) = \ex[G_t|S_t=s, \pi]$, which can be shown to be independent of $t$.
	
	In this paper, we restrict attention to policies computed by a network of $L$ stochastic units. The following definitions hold for any time $t>0$. Let $H^{(l)}_t$ denote the activation value of the unit $l \in \{1, 2, ..., L\}$ at time $t$, which is a one-dimensional discrete or continuous random variable. We also let $H^{(0)}_t= S_t$ and $A_t=H^{(L)}_t$. We call unit $l$, where $1 \leq l \leq L-1$, a hidden unit and unit $L$ the output unit. For any $1 \leq l \leq L$, the distribution of $H^{(l)}_t$ conditional on $H^{(0:l-1)}_t$ is given by $\pi_l: (\mathcal{S} \times \mathbb{R}^{l-1}) \times \mathbb{R} \rightarrow [0,1]$, such that $ \pr(H^{(l)}_t=h^{(l)}|H^{(0:l-1)}_t=h^{(0:l-1)};\mathbf{w}^{(l)})=\pi_l(h^{(0:l-1)}, h^{(l)}; \mathbf{w}^{(l)})$, where $\mathbf{w}^{(l)}$ is the parameter of unit $l$. To sample an action $A_t$ from the network given $S_t$, we iteratively sample $H^{(l)}_t \sim  \pi_l(H^{(0:l-1)}_t, \cdot ; \mathbf{w}^{(l)})$ from $l=1$ to $L$. In other words, the network is a feedforward network with units ordered and connections restricted to higher-number units. This formulation is a generalization of multi-layer networks of stochastic units since multi-layer networks are the special case of units being arranged in layers and connections between non-adjacent layers being set to zero. We mostly focus on multi-layer networks in this paper.
	
	We assume that for all $1 \leq l \leq L$, $\pi_l(h^{(0:l-1)}, h^{(l)}; \mathbf{w}^{(l)})$ can be expressed as a differentiable function $f_l(\mathbf{w}^{(l)} \cdot h^{(0:l-1)}, h^{(l)})$, where $\cdot$ denotes the dot product, and call the vector $\mathbf{w}^{(l)}$ the \emph{incoming weight} of unit $l$. We denote the weight connecting from unit $k$ to $l$ (where $k<l$) as $\mathbf{w}^{(l)}_{(k)}$, and call the vector $\mathbf{v}^{(k)} = [\mathbf{w}^{(k+1)}_{(k)}, \mathbf{w}^{(k+2)}_{(k)}, ..., \mathbf{w}^{(L)}_{(k)}]^T$ the \emph{outgoing weight} of unit $k$. 
	
	Though we restrict our attention to a network with a single output unit, the algorithm we present can be generalized to a network with multiple output units easily by only replacing the rewards to all hidden units by the change in the norm of their outgoing weights as in (\ref{eq:1}).
	
	The case we consider here is one in which all the units of the network implement an RL algorithm and share the same reward signal. These networks can therefore be considered to be \emph{teams of agents}, which have also been called \emph{coagent networks} \cite{thomas2011policy}; agents here refer to \emph{RL agents} \cite{sutton2018reinforcement}.
	
	We denote $\vert \vert \mathbf{x} \vert \vert^p_p$ as the $p$-norm of vector $\mathbf{x}$ to the power of $p$, and $x^{(m:n)}$ as $\{x^m, x^{m+1}, ..., x^{n}\}$.
	
	\section{Algorithm}
	
	\begin{figure*}[ht]
		\advance\leftskip4cm
		\includegraphics[width=0.75\textwidth]{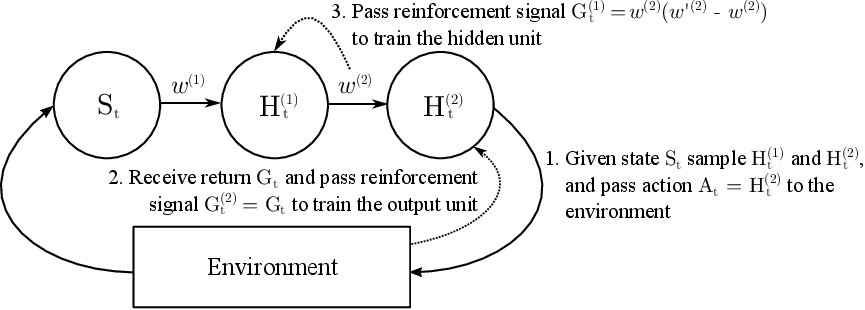}
		\caption{Illustration of a network with two units trained by Weight Maximization. See text for explanation. 	
		}%
		\label{fig:0}%
	\end{figure*}
	
	The gradient of return at time $t$ with respect to $\mathbf{w}^{(l)}$, where $1 \leq l \leq L$, can be estimated by REINFORCE, also known as the likelihood ratio estimator:
	\begin{align}
		&\nabla_{\mathbf{w}^{(l)}} \ex[G_t] = \sum_{k=t}^\infty \gamma^{(k-t)} \ex[G_k\nabla_{\mathbf{w}^{(l)}}\log \pr (A_k|S_k)]. \label{eq:0} 
	\end{align}
	We can show that the terms in the summation of (\ref{eq:0}) also equals the expectation of the update given by REINFORCE applied to each hidden unit with the same  reinforcement signal $G_t$:
	\begin{theorem}\label{thm:1}
		Let the policy be a network of stochastic units as defined above. For all $t >0$ and $1 \leq l \leq L$,
		\begin{align} &\ex[G_t\nabla_{\mathbf{w}^{(l)}}\log \pr(A_t|S_t)] \nonumber \\ 
			= &\ex[G_t\nabla_{\mathbf{w}^{(l)}} \log \pi_l(H^{(0:l-1)}_t, H^{(l)}_t; \mathbf{w}^{(l)}) ].
		\end{align}
	\end{theorem}
	See \citet{williams1992simple} for the proof. This shows that we can apply REINFORCE to each unit of the network, and the learning rule still gives an unbiased estimate of the gradient of the return. Therefore, denoting $\alpha$ as the step size, we can update parameters by the following stochastic gradient ascent rule:
	\begin{equation}
		\mathbf{w}^{(l)} \leftarrow \mathbf{w}^{(l)} + \alpha G_t\nabla_{\mathbf{w}^{(l)}} \log \pi_l(H^{(0:l-1)}_t, H^{(l)}_t; \mathbf{w}^{(l)}). \label{eq:0.5}
	\end{equation}
	However, this learning rule suffers from high variance since a single reinforcement signal ($G_t$) is used to evaluate the collective action of all units. In other words, the signal $G_t$ has a weak correlation with the gradient $\nabla_{\mathbf{w}^{(l)}} \log \pi_l(H^{(0:l-1)}_t, H^{(l)}_t; \mathbf{w}^{(l)})$, making the multiplication of these two terms have a high variance. To reduce this variance, we propose to replace this signal to each unit $l$ by: 
	\begin{equation}
		G^{(l)}_t = 
		\begin{cases}
			\mathbf{v}^{(l)} \cdot \Delta \mathbf{v}^{(l)}_t &\text{ for } l \in \{1, 2, ..., L-1\}, \\
			G_t &\text{ for } l = L.
		\end{cases}  \label{eq:1}
	\end{equation}
	where $\cdot$ denotes the dot product, and $\Delta \mathbf{v}^{(l)}_t$ is the change of the outgoing weight, $\mathbf{v}^{(l)}$, resulting from the update of the outgoing weight at time $t$. For the output unit, we let $G^{(L)}_t = G_t$; that is, the output unit is still maximizing the return from the environment.
	
	With the new reinforcement signal $G^{(l)}_t$, each hidden unit is approximately maximizing the $L^2$ norm of its outgoing weight. To see this, consider the change in the $L^2$ norm of the outgoing weight for hidden unit $l$ at time $t$:
	
	\begin{align}
		& \vert \vert  \mathbf{v}^{(l)} + \Delta \mathbf{v}^{(l)}_t \vert \vert^2_2 - \vert \vert \mathbf{v}^{(l)} \vert \vert^2_2  \\
		=& 2 \mathbf{v}^{(l)} \cdot \Delta \mathbf{v}^{(l)}_t  + \vert \vert \Delta \mathbf{v}^{(l)}_t \vert \vert^2_2. \label{eq:2}
	\end{align}
	
	We observe that $G^{(l)}_t$ is proportional to (\ref{eq:2}) except for the term $\vert \vert \Delta \mathbf{v}^{(l)}_t \vert \vert^2_2$. We choose to ignore this term in the reinforcement signal since this term is $\mathcal{O}(\alpha^2)$ while $2 \mathbf{v}^{(l)} \cdot \Delta \mathbf{v}^{(l)}_t = \mathcal{O}(\alpha)$. If the step size $\alpha$ is very small as in typical experiments, then $\vert \vert \Delta \mathbf{v}^{(l)}_t \vert \vert^2_2$ is also negligible and does not affect experimental results. However, by adjusting $\alpha$, this term can be made arbitrarily small or large, and so we choose to remove it completely instead. This removal is necessary to arrive Theorem \ref{thm:2}. 
	
	
	The motivation of using the change in the norm of a unit's outgoing weight as a reinforcement signal in (\ref{eq:1}) is based on the idea that the norm of a unit's outgoing weight roughly reflects the contribution of the unit in the network. For example, if the hidden unit's output is useful in guiding action, then the output unit will learn a large weight associated with it. Conversely, if the hidden agent is outputting random noise, then the output unit will learn a zero weight associated with it. With the new reinforcement signal, each unit gets a different local reward that evaluates its own action instead of the entire team's action, thus allowing efficient structural credit assignment. This idea of measuring the worth of a unit by its outgoing weight's norm has a long history \citep{uhr1961pattern, klopf1969evolutionary, selfridge1988pandemonium}.
	
	With the new reinforcement signal, the learning rule at time $t$ becomes:
	\begin{align}
		\mathbf{w}^{(l)} &\leftarrow \mathbf{w}^{(l)} + \Delta \mathbf{w}^{(l)}_t, \label{eq:2.4}  \\
		\Delta \mathbf{w}^{(l)}_t &= 
		\alpha G^{(l)}_t \nabla_{\mathbf{w}^{(l)}} \log \pi_l(H^{(0:l-1)}_t, H^{(l)}_t; \mathbf{w}^{(l)}), \label{eq:2.5} 
	\end{align}	
	where $1 \leq l \leq L$. Note that the only difference of the above learning rule with (\ref{eq:0.5}) is the change of the reinforcement signals to hidden units. We call this new learning method \emph{Weight Maximization}. To apply the learning rule, we compute $\Delta \mathbf{w}^{(l)}_t$ iteratively from $l=L$ to $1$.  The pseudo-code can be found in Algorithm 1 of Appendix B. The computational cost of Weight Maximization is the same as backpropagation since it is linear in the number of layers.
	
	To illustrate the algorithm, we consider an example of a simple network with only two units (one hidden unit and one output unit), as shown in Figure \ref{fig:0}.  For every time step $t$, the algorithm performs the following steps:
	\begin{enumerate}
		\item Given state $S_t$, we sample the activation value of hidden unit $H^{(1)}_t \sim \pi_1(S_t, \cdot ; {w}^{(1)})$. For example, if the unit is a Bernoulli-logistic unit and the state $S_t$ is one-dimensional, then $\pr(H^{(1)}_t=1|S_t) = \sigma(w^{(1)}S_t)$ and $\pr(H^{(1)}_t=0|S_t) = 1-\sigma(w^{(1)}S_t),$ where $\sigma$ is the sigmoid function. We sample output unit $H^{(2)}_t \sim \pi_2(H^{(1)}_t, \cdot ; {w}^{(2)})$ similarly, and pass action $A_t = H^{(2)}_t$ to the environment. 
		\item After receiving return $G_t$ from the environment (which is only known after the end of episode, but can be replaced with TD error for online learning), we use it to train the output unit by REINFORCE: $w'^{(2)} = w^{(2)} + \alpha G^{(2)}_t \nabla_{w^{(2)}} \log \pi_2(H^{(1)}_t, H^{(2)}_t ; {w}^{(2)}) $, where $\alpha$ is the step size and $G^{(2)}_t = G_t$. 
		\item We compute the reinforcement signal to the hidden unit by $G^{(1)}_t = v^{(1)} \Delta v^{(1)} = w^{(2)} (w'^{(2)} - w^{(2)})$, which  is used to train the hidden unit by REINFORCE: $w'^{(1)} = w^{(1)} + \alpha G^{(1)}_t \nabla_{w^{(1)}} \log \pi_1(S_t, H^{(1)}_t ; {w}^{(1)}) $.
	\end{enumerate}
	
	In the following, we discuss the theoretical properties of Weight Maximization.
	
	
	
	
	\subsection{Goal Alignment Condition} \label{sec:g}
	In this section, we address an important question: Under which situations is the goal of the hidden units and the whole network aligned? To simplify the discussion, we only consider single-time-step MDPs in this section and drop the subscript $t$, but the theorems here can be generalized to multiple-time-step MDPs.
	
	To understand when the hidden units are maximizing the global reward, we analyze the gradient followed by the learning rule of the hidden units. First, the output unit is maximizing the return $G$ as a result of Theorem 1 and REINFORCE:
	\begin{align}
		\ex [\Delta \mathbf{w}^{(L)}] \propto \nabla_{\mathbf{w}^{(L)}} \ex[G] \label{eq:3}.
	\end{align}
	Then consider the learning rule of unit $L-1$, which is maximizing $G^{(L-1)}$:
	\begin{align}
		\ex [\Delta \mathbf{w}^{(L-1)}] &\propto \nabla_{\mathbf{w}^{(L-1)}} \ex[G^{(L-1)}]\\
		&= \nabla_{\mathbf{w}^{(L-1)}} \ex[\mathbf{v}^{(L-1)} \cdot \Delta {\mathbf{v}^{(L-1)}} \cdot ]\\
		&\propto \nabla_{\mathbf{w}^{(L-1)}} (\mathbf{v}^{(L-1)} \cdot \nabla_{\mathbf{v}^{(L-1)}} \ex[G]  ). \label{eq:4}
	\end{align}	
	The last line is due to the fact that $\Delta \mathbf{v}^{(L-1)}$ is an entry in the vector $\Delta \mathbf{w}^{(L)}$, and so we can substitute the expectation of it with (\ref{eq:3}). We can continue the same process to derive the formulas of $\ex [\Delta \mathbf{w}^{(l)}]$ for $l=L-2, L-3, ..., 1$. This shows that the learning rule of hidden units is related to high-order cross partial derivatives of $\ex[G]$ instead of the first-order derivative. To have the goal of units and the network aligned, it is sufficient and necessary that the cross partial derivatives are the same as the first-order derivative. Formally,
	
	\begin{lemma} \label{lemma:1}	
		Let the policy be a network of stochastic units as defined above, and $\Delta \mathbf{w}^{(l)}$ be defined by (\ref{eq:2.5}). Then
		$\ex [\Delta \mathbf{w}^{(l)}] \propto  \nabla_{\mathbf{w}^{(l)}} \ex[G]$ for all $1 \leq l \leq L-1$ if and only if $\nabla_{\mathbf{w}^{(l)}} (\mathbf{v}^{(l)} \cdot \nabla_{\mathbf{v}^{(l)}} \ex[G] ) \propto \nabla_{\mathbf{w}^{(l)}} \ex[G]$ for all $1 \leq l \leq L-1$.
	\end{lemma}
	
	The proof can be found in Appendix A.1. Therefore, we define the \emph{goal alignment condition}, which is a sufficient condition that the hidden units are also maximizing the global return when applying Weight Maximization, by:
	\begin{definition}
		Let the policy be a network of stochastic units as defined above. We say that the network has satisfied the goal alignment condition in an MDP if for all $1 \leq l \leq L-1$, 
		$$
		\nabla_{\mathbf{w}^{(l)}} (\mathbf{v}^{(l)} \cdot \nabla_{\mathbf{v}^{(l)}} \ex[G] ) \propto \nabla_{\mathbf{w}^{(l)}} \ex[G].$$
	\end{definition}
	
	Except in special cases like a piecewise linear network with a piecewise linear reward function, this goal alignment condition does not hold exactly. However, the goal alignment condition holds approximately for all networks without extra assumptions:
	
	\begin{theorem}	 \label{thm:2}
		Let the policy be a network of stochastic units as defined above. For all $1 \leq l \leq L-1$, 
		$$
		\nabla_{\mathbf{w}^{(l)}} (\mathbf{v}^{(l)} \cdot \nabla_{\mathbf{v}^{(l)}} \ex[G] ) = \nabla_{\mathbf{w}^{(l)}} \ex[G] + \mathcal{O}({\vert\vert \mathbf{v}^{(l)} \vert\vert^2_2}).$$
	\end{theorem}
	
	The proof can be found in Appendix A.2. Therefore, the learning rule of hidden unit $l$ is only approximately following the gradient of return in expectation with an error of $ \sum_{k=l+1}^L \mathcal{O}({\vert\vert \mathbf{w}^{(k)}\vert\vert^2_2})$, since the error accumulates across units. In other words, the bias associated with the learning rule scales with the $L^2$ norm of the unit's outgoing weight. It is interesting to see that the more `successful' a unit is (measured by the norm of its outgoing weight), the more severe is the problem of goal misalignment. 
	
	To combat the problem of goal misalignment, we suggest adding $L^2$ regularization, or weight decay, which can be seen as a soft constraint on the $L^2$ norm of weights \cite{goodfellow2016deep} and thus prevents weights from having a large magnitude. 
	
	By replacing the rewards to the hidden units with (\ref{eq:1}) and adding weight regularization, the original cooperative game is turned into a \emph{competitive game}. A competitive game refers to the scenario where each agent is receiving different rewards \cite{sutton2018reinforcement}. With weight regularization, the upper layer has a `limited' norm of weight to allocate due to the soft constraint, and so the units in the lower layer have to compete for the limited resources. In other words, units want to maximize their outgoing weights' norm, but the outgoing units want to minimize their incoming weights' norm, and therefore competition exists between units.
	
	\begin{table*}[ht]
		\centering
		\begin{tabular}{lcccccccccccccc}
			\toprule
			& \multicolumn{2}{c}{Multiplexer} & \multicolumn{2}{c}{CartPole} & \multicolumn{2}{c}{Acrobot} & \multicolumn{2}{c}{LunarLander} \\
			\cmidrule(r){2-3}  \cmidrule(r){4-5} \cmidrule(r){6-7} \cmidrule(r){8-9}  
			& Mean & Std. & Mean & Std. & Mean & Std. & Mean & Std.  \\
			\midrule
			Weight Max &  0.81 & 0.01 & 390.38 & 43.25 & -97.05 & 2.90 & 111.23 & 16.58\\
			Weight Max w/ traces & n.a. & n.a. & 373.11 & 17.86 & -105.00 & 4.38 & 39.04 & 14.39\\		
			REINFORCE & 0.29 & 0.01 & 163.92 & 64.43 & -134.15 & 8.87 & -94.23  & 19.03 \\					
			REINFORCE with \citet{thomas2011policy} & 0.28 & 0.02 & 363.71 & 24.07 & -112.26 & 8.85 & -66.19 & 67.14  \\		
			STE Backprop & 0.76 & 0.01 & 420.73 & 18.30 & -98.55 &  5.05 & 47.53 & 35.04\\								
			Backprop & 0.84 & 0.01 & 411.82 &  25.98 &  -91.82 & 5.24 & 71.77  & 21.02 \\											
			\bottomrule
		\end{tabular}
		\caption{Average return over all episodes.}
		\label{table:1}
	\end{table*}
	
	\subsection{Weight Maximization with Eligibility Traces}
	
	Though Weight Maximization does not require a separate feedback pathway, the learning of a hidden unit needs to wait for the outgoing weight to finish updating; in contrast, in biological neurons, the change of synaptic strength has a slower timescale than the activation of neurons \cite{nicoll2017brief}, making it difficult to be used as an immediate feedback signal. Besides different timescales, the change in synaptic strength of a biological neuron is the result of the neuron's activity within a time interval instead of a single discrete time step. For example, multiple pairs of spikes are required to induce noticeable change in synaptic weights in spike-timing-dependent plasticity (STDP) experiments \cite{citri2008synaptic, gerstner2014neuronal}. In short, to be closer to biologically-observed plasticity rules, the effect of a unit's action on the outgoing weight should be slow and long-lasting, instead of immediate and precise as assumed in Weight Maximization. We propose to use eligibility traces to solve both issues.
	
	
	
	In the following discussion, we only consider a multi-layer network of stochastic units consisting of $M \geq 1$ layers. The $L$ units in the network are arranged into a multi-layer structure such that the weights connecting units on non-adjacent layers are frozen to zero. We also denote $d(l)$ as the layer in which unit $l$ resides and assume the last layer only contains the output unit.
	
	Assume each unit requires a single time step to compute the weight update. That is, the weight update of unit $l$ at time $t$, denoted by $\Delta \mathbf{w}^{(l)}_{t}$, is based on $G^{(l)}_{t-1}$ (defined in (\ref{eq:1})) and $H^{(l)}_{t-1}$, the reward and action at the previous time step. Since $\Delta \mathbf{w}^{(l)}_{t}$ does not depend on the weight change at the same time step, we can compute it for all layers in parallel. However, the reward for the hidden layer $m$ is lagging behind by $M-m$ time steps, as it takes a single time step for each of the $M-m$ upper layers to compute their weight updates. Therefore, the first issue can be seen as the problem of delayed reward - the action of a hidden unit on layer $m$ at time $t$ affects its rewards at time $t+M-m$, but not before.
	
	For the second issue, consider the scenario where the effect of a unit's action is long-lasting on the change of its outgoing weight. For example, if the output unit learns with decaying eligibility traces \cite{sutton2018reinforcement}, then all $\Delta \mathbf{w}^{(L)}_{t+1}, \Delta \mathbf{w}^{(L)}_{t+2}, ...$ depends on the value of layer $M-1$ at time $t$, though the dependence decays with time. In other words, the action of layer $M-1$ at time $t$ will affect the change in its outgoing weight at and after time $t+1$. We can continue the discussion for layer $M-2, M-3, ..., 1$. From this perspective, the second issue can again be seen as the problem of delayed reward - the action of a hidden unit on layer $m$ at time $t$ affects all its rewards at and after time $t+M-m$, but not before.
	
	The problem of delayed reward is well studied in RL, and one prominent and one biologically plausible solution is eligibility traces. We suggest using the following decay function $\lambda^l(t): \mathbb{Z} \rightarrow [0,1]$ for the unit $l$:	
	\begin{align}		
		\lambda^l(t) = 	
		\begin{cases}
			0 &\text{ for } t \leq M-d(l)-1, \\
			(1 - \lambda) \lambda^{t - (M-d(l))} &\text{ else, }
		\end{cases}  
	\end{align}
	where $\lambda \in [0,1]$ is the decay rate. Therefore, $\lambda^l(t)$ is the exponentially decaying trace but shifted by $M-d(l)$ time steps, since the action of unit $l$ does not affect the reward in the next $M-d(l)$ time steps. With this decay function, we can generalize Weight Maximization to use eligibility traces. The learning rule of \emph{Weight Maximization with eligibility traces} for a multi-layer network of stochastic units at time $t$ is given by :    
	\begin{align}
		\mathbf{w}^{(l)} \leftarrow& \mathbf{w}^{(l)} + \Delta \mathbf{w}^{(l)}_{t}, \label{eq:12.5}  \\
		\Delta \mathbf{w}^{(l)}_{t} =& \alpha \delta^{(l)}_{t-1} \mathbf{z}^{(l)}_{t-1},  \label{eq:18.5} \\
		\delta^{(l)}_t =& 
		\begin{cases}
			\mathbf{v}^{(l)} \cdot \Delta \mathbf{v}^{(l)}_t &\text{ for } l \leq L-1, \\
			R_t + \gamma V^\pi(S_{t+1}) - V^\pi(S_{t}) &\text{ for } l = L,
		\end{cases}  \label{eq:19} \\
		\mathbf{z}^{(l)}_t =& \sum_{k=0}^{t-1} \lambda^l(k) \gamma^{k-(M-d(l))} \nonumber \\ &\nabla_{\mathbf{w}^{(l)}} \log \pi_l(H^{(0:l-1)}_{t-k}, H^{(l)}_{t-k}; \mathbf{w}^{(l)}),
	\end{align}	
	where $\alpha$ denotes the step size and $1 \leq l \leq L$. We also replace the return $G_t$ by TD error $R_t + \gamma V^\pi(S_{t+1}) - V^\pi(S_{t})$ to make the algorithm online. In practice, the state values $V^\pi(S_{t})$ is generally unknown, but we can estimate the state values by another network implementing a TD algorithm, which is called a critic network \citep{sutton2018reinforcement}. The pseudo-code can be found in Algorithm 2 of Appendix B.
	
	With the above modification, the effect of a unit's action on its outgoing weight is slow and long-lasting, in accordance with biologically-observed plasticity rules. Weight Maximization with eligibility traces also solves the three problems of backprop discussed in the introduction. Its learning rule is local and does not require any separate feedback pathways. The algorithm can be implemented in parallel for all layers, and there are no distinct feedforward and feedback phases for the whole network. 
	
	It should be noted that the feedback signal in backprop can also be computed based on the change of a unit's outgoing weight in some cases, eliminating the need for a separate feedback pathway. However, as discussed earlier, it is not biologically plausible to use the change of a unit's outgoing weight as an immediate feedback signal. The solution presented in this section, namely eligibility traces, can only be applied to Weight Maximization but not backprop, since backprop requires the activation values to be precisely matched with the feedback signals at the same time step. This underlines one major difference between the two algorithms.

	\section{Related Work}
	Research on solving tasks by a team of RL agents has a long history. \citet{narendra1974learning, narendra2012learning} described the stochastic learning automata, and \citet{klopf1972brain, klopf1982hedonistic} proposed the hedonistic neuron hypothesis, which conjectured that individual neurons seek to maximize their own pleasure, and the collective behavior of these neurons can yield powerful adaptive systems. \citet{barto1985pattern} and \citet{barto1985learning} introduced the $A_{R-\lambda P}$ algorithm and showed that a team of $A_{R-\lambda P}$ units could learn with a globally-broadcast reward signal. Extending this class of learning rule, \citet{williams1992simple} introduced REINFORCE, a special case of $A_{R-\lambda P}$ when $\lambda=0$, and proved that a team of agents trained with REINFORCE ascends the average reward gradient. Such a team of agents is recently called coagent networks \cite{thomas2011policy}. Theories relating to training coagent networks have been investigated \citep{thomas2011policy, kostasasynchronous, thomas2011conjugate}, and \citet{thomas2011policy} proposed a variance reduction method for training coagent networks with REINFORCE by disabling exploration randomly. \citet{chung2021map} proposed the MAP propagation algorithm, which minimizes the energy of the network before applying REINFORCE, to reduce the variance efficiently. However, in these papers, each agent in the team receives the same reward signal. In contrast, we propose that each agent maximizes the norm of its outgoing weight instead of the same reward signal, which turns the problem from a \emph{cooperative game} into a \emph{competitive game}. \citet{zhang2019multi} reviewed recent development in the wider field of multi-agent RL.
	
	Measuring the worth of a unit by the norm of its outgoing weight has been proposed  \citep{uhr1961pattern, klopf1969evolutionary, selfridge1988pandemonium}. In these papers, hidden units with small outgoing weights are replaced by new hidden units with random incoming weights. These methods are thus based on the evolution approach instead of the RL approach as in Weight Maximization. \citet{anderson1986learning} proposed \emph{$A_{R-P}$ algorithm with Penalty Prediction}, which pushes units with small outgoing weights to match the incoming values. Bucket Brigade \citep{holland1985properties} algorithm uses the strength of a classifier to do credit assignment, but with a winner-take-all selection and in classifier systems. 
	
	In addition, there is a large literature on methods for training a network of stochastic units, and a review can be found in \citet{weber2019credit}. STE backprop \cite{bengio2013estimating} is a practical method of training a network of stochastic discrete units. Though STE backprop does not follow the gradient of the loss function, it is arguably the most effective way of training quantized ANN \cite{courbariaux2015binaryconnect, rastegari2016xnor} and
	\citet{yin2019understanding} provides some theoretical justification for STE backprop. However, STE backprop suffers the same problem with backprop regarding biological plausibility.
	
	Besides a team of agents trained by REINFORCE, many biologically plausible alternatives to backprop have been proposed. Biologically plausible learning rules based on reward prediction errors and attentional feedback have been proposed \citep{pozzi2020attention, roelfsema2005attention, rombouts2015attention}; but these learning rules mostly require a non-local feedback signal. Moreover, local learning rules based on contrastive divergence or nudging the values of output units have been proposed \citep{movellan1991contrastive, hinton2002training, scellier2017equilibrium}. See \citet{lillicrap2020backpropagation} for a comprehensive review of algorithms that approximate backprop with local learning rules based on the differences in units' values. Contrary to these papers, Weight Maximization does not require any separate feedback pathways or distinct phases in learning. Also, most of these algorithms are applied in supervised or unsupervised learning tasks, while Weight Maximization is applied in RL tasks.
	
	\section{Experiments}
	
	\begin{figure*}[ht]
		\centering
		\includegraphics[width=0.9\textwidth]{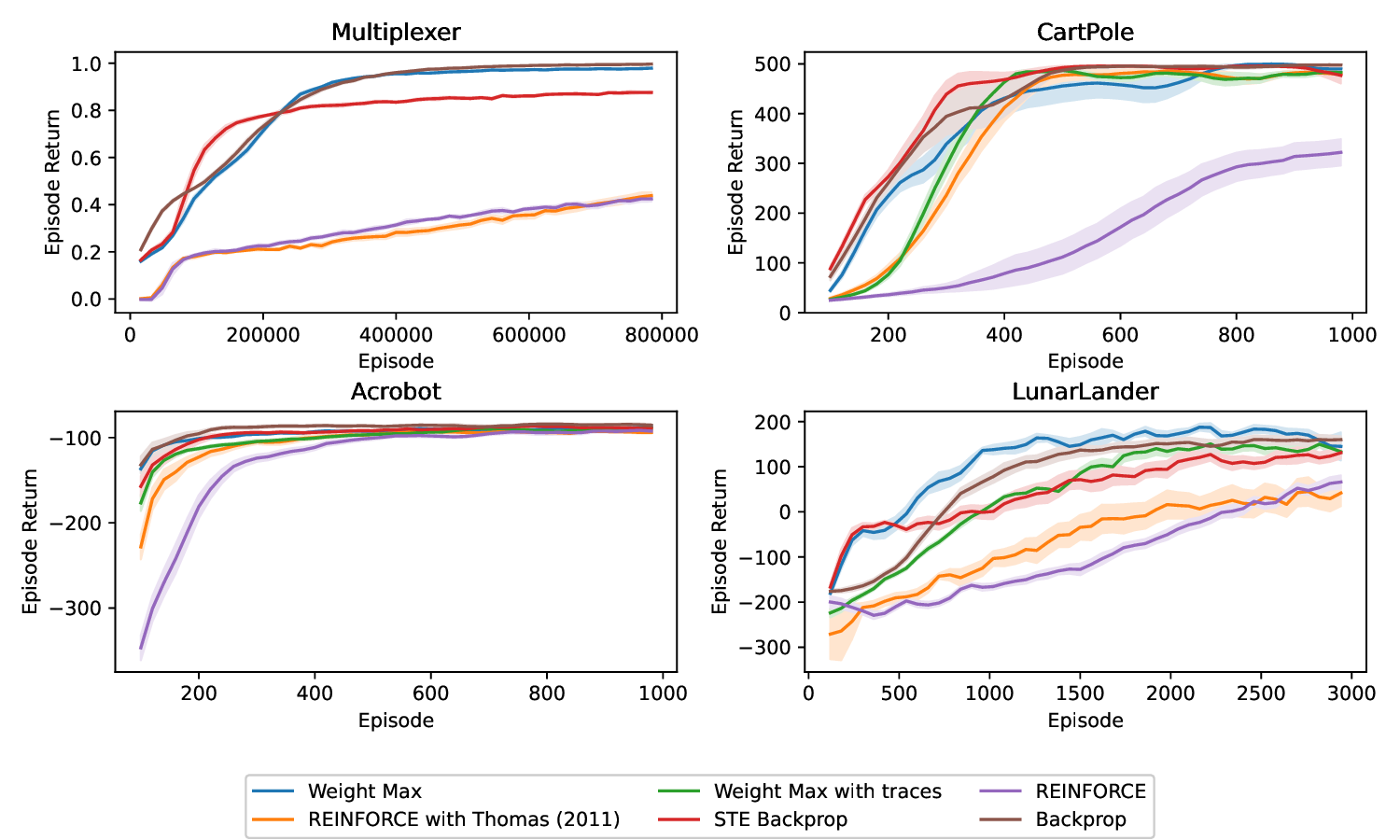}
		\caption{Episode returns in different RL tasks. Results are averaged over 10 independent runs, and shaded areas represent standard deviation over the runs. Curves are smoothed with a running average of 100 episodes (10,000 episodes for the multiplexer task).}%
		\label{fig:1}%
	\end{figure*}
	
	\begin{figure*}[h]
		\centering
		\includegraphics[width=0.9\textwidth]{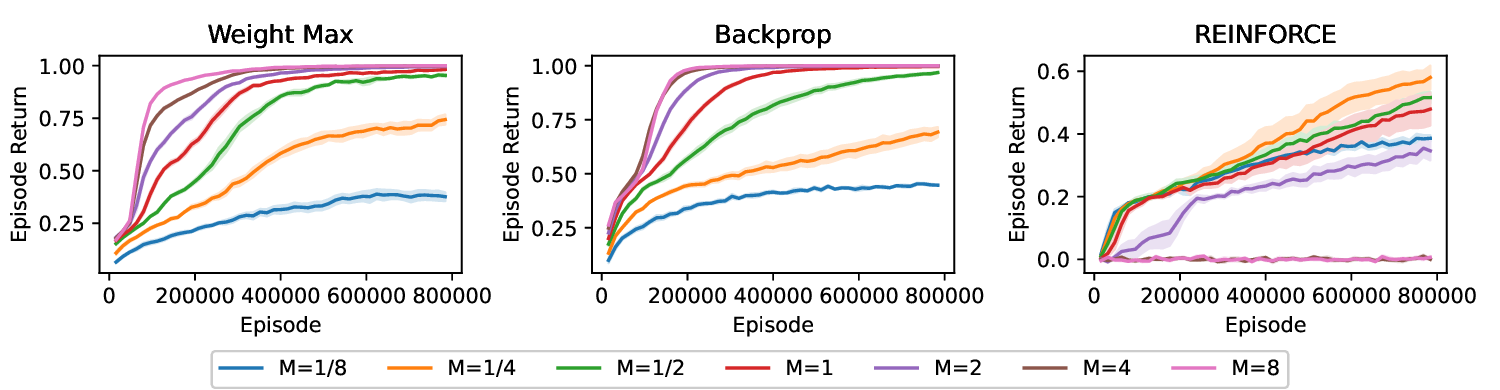}
		\caption{Episode returns in the multiplexer task with a varying number of units in the network. Results are averaged over 10 independent runs, and shaded areas represent standard deviation over the runs. Curves are smoothed with a running average of 10,000 episodes.}%
		\label{fig:2}%
	\end{figure*}
	
	\begin{figure*}[h]
		\centering
		\includegraphics[width=0.9\textwidth]{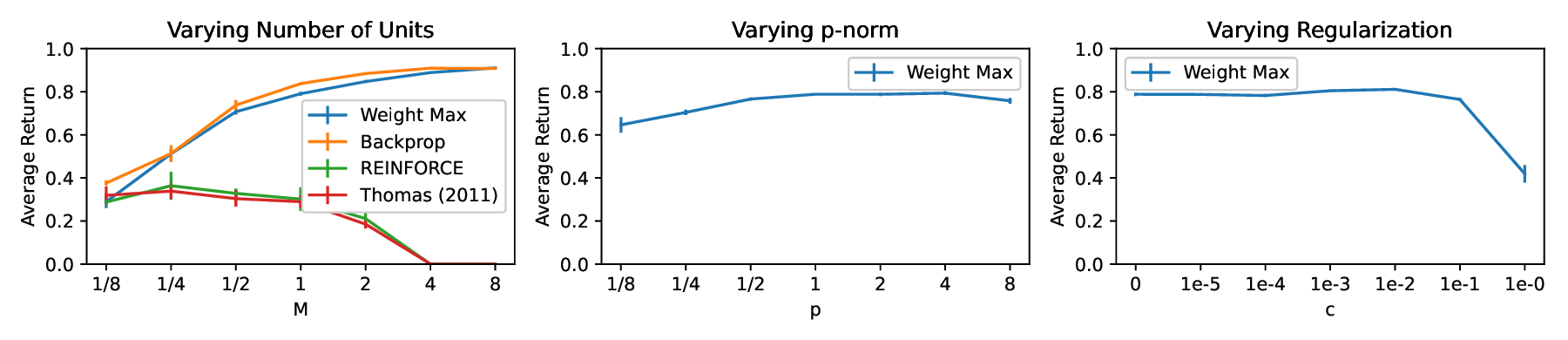}
		\caption{Average returns of all episodes in the multiplexer task with different hyperparameters. Results are averaged over 10 independent runs.}%
		\label{fig:3}%
	\end{figure*}
	
	We applied our algorithms to four RL tasks: multiplexer, CartPole, Acrobot, and LunarLander. Details of the tasks can be found in Appendix C. We tested two variants of Weight Maximization: i. Weight Maximization, ii. Weight Maximization with Eligibility Traces. For both variants, we used Algorithm 2 in Appendix B (i. corresponds to the case $\lambda=0$). We did not test Weight Maximization with Eligibility Traces on the multiplexer task since the task only has a single time step. 
	
	All networks considered have the same architecture: a three-layer network of stochastic units, with the first hidden layer having 64 units, the second hidden layer having 32 units, and the output layer being a softmax layer. All hidden units are Bernoulli-logistic units, i.e. $\pi_l(h^{(0:l-1)}, h^{(l)}, \mathbf{w}^{(l)}) = \sigma(\mathbf{w}^{(l)} \cdot h^{(0:l-1)})^{h^{(l)}} (1-\sigma(\mathbf{w}^{(l)} \cdot h^{(0:l-1)}))^{1-h^{(l)}}$, where $\sigma$ is the sigmoid function.
	
	We consider four baselines to train hidden units: i. REINFORCE, ii. REINFORCE with the variance reduction method proposed by \citet{thomas2011policy}, iii. STE backprop \cite{bengio2013estimating} and iv. backprop. For i. to iii., the networks are the same as the one used in Weight Maximization. For iv., the Bernoulli-logistic units are replaced with deterministic rectified linear units (ReLUs) so the hidden units can be trained by backprop. In all baselines, the output unit was trained by REINFORCE, and we used eligibility traces. Thus, all learning methods in the experiments are variants of Actor-Critic with Eligibility Traces (episodic) \cite{sutton2018reinforcement} with different methods of accumulating trace (for the baselines) or different reward signals to each unit (for Weight Maximization). For the critic networks in all experiments, we used a three-layer ANN trained by backprop. Other experiments' details and choice of hyperparameters can be found in Appendix C. The code is available at \url{https://github.com/stephen-chung-mh/weight_max}.
	
	It should be noted that there are other algorithms besides actor-critic networks that can solve the tasks more efficiently, and the three-layer architecture is also chosen prior to experiments, which is not the optimal architecture since a shallower network performs better for simple tasks. The experiments only aim to compare different training methods for hidden units in a multi-layer actor network.
	
	The average return over ten independent runs is shown in Fig \ref{fig:1}.  The mean and standard deviation of the average return can be found in Table \ref{table:1}. We observe that both variants of Weight Maximization have a significantly better performance than REINFORCE, suggesting that Weight Maximization allows effective structural credit assignment. Also, Weight Maximization has a similar performance compared to STE backprop, indicating that it is an effective method for training discrete units. 
	
	However, compared to a network of continuous-valued units trained by backprop, Weight Maximization performed slightly worse (except for the LunarLander). This is likely due to the limitation of discrete-valued units - units can only communicate with binary values instead of real values. This view is supported by the observation that discrete-valued units trained by STE backprop also performed worse than backprop. However, discrete units have the advantages of low memory and communication costs; thus the slower learning may be compensated by a more efficient computation.
	
	We notice that adding traces to Weight Maximization does not improve the performance. This is likely due to the harder temporal credit assignment. By using traces, the change in a unit's outgoing weight at time $t$ is affected by the unit's action at time $t, t-1, t-2, ...$ instead of only time $t$, making temporal credit assignment more difficult and thus learning slower.
	
	In addition, we found that the representation learned by Weight Maximization is more statistically independent than backprop. For example, after training an agent to solve CartPole by Weight Maximization (STE backprop), the absolute correlation across units on the first and the second layer is 0.690 (0.888) and 0.640 (0.830) respectively on average. This may be explained by the dynamics of Weight Maximization - units compete with each other to be more `useful' and so have to learn different signals. 
	
	To better understand how different learning rules scale with the number of units in a network, we repeated the experiments on the multiplexer tasks with varying numbers of units in the network. Let the network have $64M$ and $32M$ units on the first and the second layer of the network respectively. The experimental results with different $M$ can be found in Figure \ref{fig:2} and Figure \ref{fig:3}.
	
	We observe that the performance of both Weight Maximization and backprop improves with a larger $M$, while the performance of REINFORCE peaks at $M=1/4$. This illustrates a critical issue of REINFORCE that renders it impractical to train large networks - as all units are independently exploring and a single reward signal is used to evaluate the collective exploration of all units, it is more difficult to assign credit to each unit with a larger network. Thus, a larger network leads to a higher variance of parameters' updates and slower learning. This issue is alleviated in Weight Maximization since each unit receives a different reward signal that is strongly correlated with the unit's own action.
	
	In Weight Maximization, we replace the reward to each unit by the change in the $L^2$ norm of the outgoing weight as shown in (\ref{eq:1}). It is also possible to use $L^p$ norm instead of $L^2$ norm by generalizing (\ref{eq:1}) (see details in Appendix C). The experimental results on the multiplexer tasks with varying $p$ can be found in Figure \ref{fig:3}. The results show that the network is also able to learn with a similar performance when using other $L^p$ norms. 
	
	We tested the effects of $L^2$ weight regularization, with $c \geq 0$ being the strength of weight regularization. That is, we add $-2c \mathbf{w}^{(l)}$ to the learning rule (\ref{eq:2.4}). The experimental results on the multiplexer tasks with varying $c$ can be found in Figure \ref{fig:3}. We observe that the performance peaks at $c=0.01$, suggesting that a small weight regularization can improve performance. This is in line with our analysis of the goal alignment condition.
	
	We note that it is inefficient to apply Weight Maximization directly to train continuous-valued units. Additional mechanisms, such as equipping each unit with a baseline, are required to match the speed of backprop and are left for future work.
	
	\section{Future Work and Conclusion}
	
	The approximate equivalence of every unit maximizing the global return in a cooperative game and every unit maximizing the norm of its outgoing weight in a competitive game offers a wide range of possible methods to train a multi-layer network efficiently besides backprop. Since each hidden unit faces a local optimization problem from the alternative perspective, learning rules besides REINFORCE can be applied to train each hidden unit. The lack of central coordination in Weight Maximization also leads to the possibility of implementing the algorithm asynchronously and efficiently with neuromorphic circuits \cite{indiveri2011neuromorphic}.
	
	
	In conclusion, we propose a novel algorithm that reduces the variance associated with training a team of agents with REINFORCE and thus significantly increases the learning speed. The proposed algorithm solves several major problems of backprop relating to biological plausibility.  We also analyze the theoretical properties of the proposed algorithm and establish the approximate equivalence of hidden units maximizing the outgoing weights and the external rewards.
	
	\section{Acknowledgment}
	We would like to thank Andrew G. Barto, who inspired this research and provided valuable insights and comments.
	
	\bibliography{citation.bib}
	\clearpage

	\appendix
	\onecolumn
	
\section{A. Proof}
\subsection{A.1 Proof of Lemma 1} \label{proof:l1}

\begin{proof}
	Assume $\ex [\Delta \mathbf{w}^{(l)}] \propto  \nabla_{\mathbf{w}^{(l)}} \ex[G]$ for all $1 \leq l \leq L-1$. Let $1 \leq l \leq L-1$. Then,
	\begin{align}
		&\nabla_{\mathbf{w}^{(l)}} \ex[G]\\
		\propto& \ex [\Delta \mathbf{w}^{(l)}] \\ 
		\propto& \ex [(\mathbf{v}^{(l)} \cdot \Delta {\mathbf{v}^{(l)}}) \nabla_{\mathbf{w}^{(l)}} \log \pi_l(H^{(0:l-1)}, H^{(l)}; \mathbf{w}^{(l)})]\\
		=& \nabla_{\mathbf{w}^{(l)}} (\mathbf{v}^{(l)} \cdot \ex [\Delta {\mathbf{v}^{(l)}} ])\\
		\propto& \nabla_{\mathbf{w}^{(l)}} (\mathbf{v}^{(l)} \cdot \nabla_{\mathbf{v}^{(l)}} \ex[G]).
	\end{align}
	Conversely, assume $\nabla_{\mathbf{w}^{(l)}} (\mathbf{v}^{(l)} \cdot \nabla_{\mathbf{v}^{(l)}} \ex[G] )\propto \nabla_{\mathbf{w}^{(l)}} \ex[G]$ for all $1 \leq l \leq L-1$. We prove by induction on $k$: For $k=L$, $\ex [\Delta \mathbf{w}^{(k)}] \propto \nabla_{\mathbf{w}^{(k)}} \ex[G]$. Let $1 \leq k < L$ and assume $\ex [\Delta \mathbf{w}^{(l)}] \propto \nabla_{\mathbf{w}^{(l)}} \ex[G]$ for all $k+1 \leq l \leq L$.  Then,
	\begin{align}
		&\nabla_{\mathbf{w}^{(k)}} \ex[G]\\
		\propto& \nabla_{\mathbf{w}^{(k)}} (\mathbf{v}^{(k)} \cdot \nabla_{\mathbf{v}^{(k)}} \ex[G] ) \\ 
		\propto& \nabla_{\mathbf{w}^{(k)}} (\mathbf{v}^{(k)} \cdot \ex [\Delta \mathbf{v}^{(k)}]) \\
		=& \ex [(\mathbf{v}^{(k)} \cdot \Delta {\mathbf{v}^{(k)}} ) \nabla_{\mathbf{w}^{(k)}} \log \pi_l(H^{(0:k-1)}, H^{(k)}; \mathbf{w}^{(k)})]\\ 
		\propto& \ex [\Delta \mathbf{w}^{(k)}].
	\end{align}	
\end{proof}	
\subsection{A.2 Proof of Theorem 2} \label{proof:t2}
\begin{proof}
	Let $1 \leq l \leq L-1$. Then, 
	{\small  
	\begin{align}
		&\nabla_{\mathbf{w}^{(l)}} (\mathbf{v}^{(l)} \cdot \nabla_{\mathbf{v}^{(l)}} \ex[G]  )\\
		=& \nabla_{\mathbf{w}^{(l)}} ( \mathbf{v}^{(l)} \cdot\nabla_{\mathbf{v}^{(l)}} \sum_{h^{(0:l)}} \pr(H^{(0:l)}=h^{(0:l)}) \ex[G|H^{(0:l)}=h^{(0:l)}] ) \label{eq:5}\\
		=& \nabla_{\mathbf{w}^{(l)}} ( \mathbf{v}^{(l)} \cdot \nabla_{\mathbf{v}^{(l)}} \sum_{h^{(0:l)}} \pr(H^{(0:l)}=h^{(0:l)}) g(h^{(l)}\mathbf{v}^{(l)}; h^{(0:l-1)}) ) \label{eq:6}\\
		=& \nabla_{\mathbf{w}^{(l)}} ( \mathbf{v}^{(l)} \cdot \sum_{h^{(0:l)}} \pr(H^{(0:l)}=h^{(0:l)}) h^{(l)} \nabla g(h^{(l)}\mathbf{v}^{(l)}; h^{(0:l-1)}))\\
		=& \nabla_{\mathbf{w}^{(l)}} \sum_{h^{(0:l)}} \pr(H^{(0:l)}=h^{(0:l)}) (h^{(l)}\mathbf{v}^{(l)} \cdot \nabla g(h^{(l)}\mathbf{v}^{(l)}; h^{(0:l-1)})) \label{eq:7}\\
		=& \nabla_{\mathbf{w}^{(l)}}  \sum_{h^{(0:l)}} \pr(H^{(0:l)}=h^{(0:l)}) (g(h^{(l)}\mathbf{v}^{(l)}; h^{(0:l-1)}) - g(\mathbf{0}; h^{(0:l-1)}) + \mathcal{O}({\vert\vert \mathbf{v}^{(l)} \vert\vert^2_2})) \label{eq:8}\\
		=& \nabla_{\mathbf{w}^{(l)}}  \sum_{h^{(0:l)}} \pr(H^{(0:l)}=h^{(0:l)}) g(h^{(l)}\mathbf{v}^{(l)}; h^{(0:l-1)})  + \mathcal{O}({\vert\vert \mathbf{v}^{(l)} \vert\vert^2_2}) \label{eq:8.1} \\	
		=& \nabla_{\mathbf{w}^{(l)}}  \sum_{h^{(0:l)}} \pr(H^{(0:l)}=h^{(0:l)}) \ex[G|H^{(0:l)}=h^{(0:l)}]  + \mathcal{O}({\vert\vert \mathbf{v}^{(l)} \vert\vert^2_2}) \label{eq:8.2} \\
		=& \nabla_{\mathbf{w}^{(l)}}  \ex[G]  + \mathcal{O}({\vert\vert \mathbf{v}^{(l)} \vert\vert^2_2}).
	\end{align}	}%
	(\ref{eq:5}) to (\ref{eq:6}) uses the fact that $\ex[G|H^{(0:l)}=h^{(0:l)}]$ can be expressed as a differentiable function $g(h^{(l)}\mathbf{v}^{(l)}; h^{(0:l-1)})$:
	{\small  
	\begin{align}
		&\ex[G|H^{(0:l)}=h^{(0:l)}] \\
		=& \sum_{h^{(l+1:L)}} \ex[G|S=h^{(0)}, A=h^{(L)}] \pi_L(h^{(0:L-1); }, h^{(L)};\mathbf{w}^{(L)})   \pi_{L-1}(h^{(0:L-2)}, h^{(L-1)};\mathbf{w}^{(L-1)}) ... \pi_{l+1}(h^{(0:l)}, h^{(l+1);\mathbf{w}^{(l+1)}}) \label{eq:9}\\
		=& \sum_{h^{(l+1:L)}} \ex[G|S=h^{(0)}, A=h^{(L)}]  f_L(\mathbf{w}^{(L)} \cdot h^{(0:L-1)}, h^{(L)})   f_{L-1}(\mathbf{w}^{(L-1)} \cdot h^{(0:L-2)}, h^{(L-1)})  ... f_{l+1}(\mathbf{w}^{(l+1)} \cdot h^{(0:l)}, h^{(l+1)})  \label{eq:10} \\
		=& g(h^{(l)}\mathbf{v}^{(l)}; h^{(0:l-1)}),
	\end{align}}%
	where (\ref{eq:9}) to (\ref{eq:10}) uses the assumption that for all $l$, $\pi_l(h^{(0:l-1)}, h^{(l)}; \mathbf{w}^{(l)})$ can be expressed as a function $f_l(\mathbf{w}^{(l)} \cdot h^{(0:l-1)}, h^{(l)})$. \\
	(\ref{eq:7}) to (\ref{eq:8}) uses the Taylor’s formula: $ g(\mathbf{0}; h^{(0:l)}) = g(h^{(l)}\mathbf{v}^{(l)}; h^{(0:l-1)}) - h^{(l)}\mathbf{v}^{(l)} \cdot \nabla g(h^{(l)}\mathbf{v}^{(l)}; h^{(0:l-1)}) +  \mathcal{O}({\vert\vert \mathbf{v}^{(l)} \vert\vert^2_2})$.\\
	(\ref{eq:8}) to (\ref{eq:8.1}) uses the fact that $g(\mathbf{0}; h^{(0:l-1)})$ does not depend on $h^{(l)}$ thus the derivative of its expectation w.r.t. $\mathbf{w}^{(l)}$ is $0$.		
\end{proof}	

\section{B. Algorithms} \label{sec:b}

\begin{algorithm}
	\textbf{Input:} differentiable policy function: $\pi_l(h^{(0:l-1)}, h^{(l)}; \mathbf{w}^{(l)})$  for $l \in \{1, 2, ..., L\}$\;
	\textbf{Algorithm Parameter:} step size $\alpha >0$ ; discount rate $\gamma \in [0,1]$\;
	\textbf{Initialize policy parameter:} $\mathbf{w}^{(l)}$ for {$l \in \{1, 2, ..., L\}$}\;
	
	\SetKwProg{lf}{Loop forever (for each episode):}{}{}
	\SetKwProg{lw}{Loop for each step of the episode, $t=0, 1, ..., T-1$:}{}{}
	\lf{}{  
		Generate an episode $S_0, H_0, R_1, ..., S_{T-1}, H_{T-1}, R_T$ following $\pi_l(\cdot, \cdot; \mathbf{w}^{(l)})$ for $l \in \{1, 2, ..., L\}$\;  
		\lw{}{
			$G^{(L)} \leftarrow \sum_{k=t+1}^T \gamma^{k-t-1}R_k$ \;     
			\For{$l=L, L-1, ..., 1$}{
				$\Delta \mathbf{w}^{(l)}  \leftarrow G^{(l)} \nabla_{\mathbf{w}^{(l)}} \log \pi_{l}(H^{(0:l-1)}_t, H^{(l)}_t; \mathbf{w}^{(l)})$ \;
				$\mathbf{w}^{(l)} \leftarrow \mathbf{w}^{(l)} + \alpha \Delta \mathbf{w}^{(l)} $\;
				$G^{(l-1)} \leftarrow \sum_{k=l}^L \mathbf{w}^{(k)}_{(l-1)} \Delta \mathbf{w}^{(k)}_{(l-1)}$  \;
	}}}
	
	\caption{Weight Maximization} \label{alg:1}
\end{algorithm}

\begin{algorithm}
	\textbf{Input:} differentiable policy function: $\pi_m(h^{(m-1)}, h^{(m)}; W^{(m)})$  for $m \in \{1, 2, ..., M\}$\;
	\textbf{Algorithm Parameter:} step size $\alpha >0$; trace decay rate $\lambda \in [0,1]$; discount rate  $\gamma \in [0,1]$\;
	\textbf{Initialize policy parameter:} $W^{(m)}$  for {$l \in \{1, 2, ..., M\}$}\;
	\SetKwProg{lf}{Loop forever (for each episode):}{}{} 
	\SetKwProg{lw}{Loop while $S_t$ is not terminal (for each time step $t=1,2, ...$):}{}{}
	\lf{}{
		Initialize $S_1$ (first state of episode) \;
		Initialize zero eligibility traces $\mathbf{z}^{(m)}$ and zero $\Delta W^{(m)}$ for {$m \in \{1, 2, ..., M\}$} \;
		\lw{}{
			$H^{0}_t \leftarrow S_t$ \;    
			\tcc{Sample Action}  	 
			Sample $H^{(m)}_t$ from $\pi_{m}(H^{(m-1)}_t, \cdot; W^{(m)})$  for {$m \in \{1, 2, ..., M\}$} \; 	
			
			\tcc{Compute $\delta^{(m)}$ and apply REINFORCE}      
			\If {\normalfont{\textbf{episode not in first time step}}}{ 
				Receive TD error $\delta$ from the critic network\;
				$\delta^{(M)}   \leftarrow \delta $\;
				$\delta^{(m)} \leftarrow (W^{(m+1)} \odot \Delta W^{(m+1)})^T \mathbf{1}$ for {$m \in \{1, 2, ..., M-1\}$}\;
				$\Delta W^{(m)} \leftarrow (\delta^{(m)} \mathbf{1}^T) \odot \mathbf{z}^{(m)}$  for {$m \in \{1, 2, ..., M\}$}\;
				$ W^{(m)} \leftarrow W^{(m)} + \alpha \Delta  W^{(m)}$	for {$m \in \{1, 2, ..., M\}$}\;
			}
			\tcc{Trace accumluation}     
			$\mathbf{z}^{(m)}  \leftarrow \gamma \lambda \mathbf{z}^{(m)} + \nabla_{W^{(m)}} \log \pi_{m}(H^{(m-1)}_{t-M+m}, H^{(m)}_{t-M+m}; W^{(m)})$ for {$m \in \{1, 2, ..., M\}$}\;	
			Take action $H^{(M)}_t$, observe $S_{t+1}, R_{t+1}$ \;
		}{}
	}{}	
	\caption{Weight Maximization with Eligibility Traces} \label{alg:3}
\end{algorithm}

\begin{table}
	\caption{The hyperparameters of Weight Maximization used in the experiments.}
	\label{table:2}
	\begin{center}
		\begin{tabular}{lcccccccc}
			\toprule[0.1ex]
			& Multiplexer & CartPole & Acrobot & LunarLander\\
			\midrule    		
			$\alpha_1$ & 1e-2 & 2e-2 & 1e-1 & 8e-2 \\	 
			$\alpha_2$ & 1e-3 & 2e-4 & 1e-3 & 8e-4\\	 
			$\alpha_3$ & 1e-4 & 2e-5 & 1e-4 & 8e-5 \\	 		
			$T$ & 1 & 1 & 1 & 2  \\	
			$\gamma$ & n.a. & .98 & .98 & .98  \\				
			\bottomrule[0.25ex]
		\end{tabular}
	\end{center}
\end{table}

The notations in Algorithm \ref{alg:3} are slightly different from the main paper: for $1 \leq m \leq M$, we let $H^{(m)}_t$ denote the values of units on layer $m$ at time $t$, which is a multivariate random variable. The distribution of $H^{(m)}_t$ conditioned on $H^{(m-1)}_t$ is given by $\pr(H^{(m)}=h^{(m)}|H^{(m-1)}=h^{(m-1)})  = \pi_m(h^{(m-1)}, h^{(m)}; W^{(m)})$, where $W^{(m)} \in \mathbb{R}^{n(m) \times n(m-1)}$ is the incoming weights for layer $m$ and $n(m)$ denotes the number of units on layer $m$.
With this new formulation, (16) can be expressed as:		
$$
\delta^{(m)} = 
\begin{cases}
	(W^{(m+1)} \odot \Delta W^{(m+1)}_t)^T \mathbf{1} &\text{ for } m \in \{1, 2, ..., M-1\}, \\
	R_t + \gamma V^\pi(S_{t+1}) - V^\pi(S_{t}) &\text{ for } m = M,
\end{cases}  
$$
where $\odot$ is elementwise multiplication and $\mathbf{1}$ is a vector of ones. Note that $\delta^{(m)} \in \mathbb{R}^{n(m)}$ for $m<M$ instead of $\mathbb{R}$ as in original notation, since there are $n(m)$ different rewards to each $n(m)$ units on layer $m$. Also, (15) can be expressed as:	
$$
\Delta W^{(m)}_t = ( \delta^{(m)}_{t-1} \mathbf{1}^T) \odot \mathbf{z}^{(m)}_{t-1}
$$
which is equivalent to multiplying each row $i$ of $\mathbf{z}^{(m)}_{t-1}$ by the entry $i$ of $\delta^{(m)}_{t-1}$.

Note that in line 17 of Algorithm \ref{alg:3}, the trace accumulation is delayed by $M-m$ time steps, and so the last $M-m$ actions of layer $m$ have to be stored in a buffer. If $t-M+m < 1$, then the gradient is replaced with $0$.

\section{C. Implementation Details of Experiments}	\label{sec:e}

In the multiplexer task, the state is sampled from all possible values of a binary vector of size $k + 2^k$ with equal probability.  The action set is $\{-1, 1\}$, and we give a reward of +1 if the action of the agent is the desired action and -1 otherwise. The desired action is given by the output of a $k$-bit multiplexer, with the input of the $k$-bit multiplexer being the state. We consider $k=4$ here, so the dimension of the state space is 20. This is similar to the 2-bit multiplexer considered by \citet{barto1985learning}. For the implementation of the remaining three tasks, we used CartPole-v1, Acrobot-v1, and LunarLander-v2 in OpenAI's Gym \cite{brockman2016openai}.

The hyperparameters of Weight Maximization used in the experiments can be found in Table \ref{table:2}. We used a different learning rate $\alpha$ for each layer of the network, and we denote the learning rate for the $l$\textsuperscript{th} layer to be $\alpha_l$. For the learning rule in line 15 of the pseudo-code, we used Adam optimizer \cite{kingma2014adam} instead, with $\beta_1=0.9$ and $\beta_2=0.999$. We did not use any batch update. We denote the temperature of the output softmax layer to be $T$. These hyperparameters were selected based on manual tuning to optimize the learning curve. We did the same manual tuning for the hyperparameters of the baseline models. For softmax layers, the last logit unit is set to be a constant zero since $n-1$ logits are sufficient to represent a distribution of a random variable with $n$ possible values.

For all learning methods with eligibility traces (including the baseline models), we used $\lambda=0.8$ except for the LunarLander task, where we used $\lambda=0.9$. The value of $\lambda$ was selected prior to the experiments, similar to the choice of the network architecture. 

We did not use critic networks for all experiments in the multiplexer task and used the reward from the environment as the reinforcement signal since there is only a single time step in the task. For the critic networks in other tasks, we used a three-layer ANN, with the same architecture as the actor network but with ReLu hidden units and a linear output unit. The critic network was trained by backprop as in Actor-Critic with Eligibility Traces (episodic) \cite{sutton2018reinforcement}.

We annealed the learning rate linearly such that the learning rate is reduced to 1e-2 and 1e-1 of the initial learning rates at 5e4 and 1e6 steps in CartPole and Acrobot respectively, and the learning rate remains unchanged afterward. We also annealed the learning rate linearly for the baseline models. We found that this can make the final performance more stable. We did not anneal the learning rate in the other two tasks.

We use the following formula to generalize Weight Maximization to use $L^p$ norm instead of $L^2$ norm (note that this generalizes (4)):

\begin{equation}
	G^{(l)}_t = 
	\begin{cases}
		(\nabla_{\mathbf{v}^{(l)}} \vert \vert \mathbf{v}^{(l)} \vert \vert ^p_p) \cdot \Delta \mathbf{v}^{(l)}_t &\text{ for } l \in \{1, 2, ..., L-1\}, \\
		G_t &\text{ for } l = L.
	\end{cases}  
\end{equation}

\end{document}